\documentclass[11pt]{article}

\usepackage[final]{acl}

\usepackage{times}
\usepackage{latexsym}

\usepackage[T1]{fontenc}

\usepackage[utf8]{inputenc}

\usepackage{microtype}
\usepackage{amsmath}

\usepackage{inconsolata}

\usepackage{graphicx}
\usepackage{url}
\usepackage{xurl}

\usepackage{multirow} 

\usepackage{enumitem}
\setlist[itemize]{leftmargin=*,topsep=2pt,itemsep=0pt,parsep=0pt,partopsep=0pt}
\setlist[enumerate]{leftmargin=*,topsep=2pt,itemsep=0pt,parsep=0pt,partopsep=0pt}

\title{AuDQ: Auditable Data Quality through Agentic Rule Discovery}

\author{
  \textbf{Jacopo Biggiogera\textsuperscript{1}} \quad
  \textbf{Nicolas Colombo\textsuperscript{1}} \quad
  \textbf{Krati Saxena\textsuperscript{2}} \\
  \textbf{Bhaskarjit Sarmah\textsuperscript{2}} \quad
  \textbf{Stefano Pasquali\textsuperscript{3}} \\[3pt]
  \textsuperscript{1}Domyn, Milan, Italy \quad
  \textsuperscript{2}Domyn, Gurugram, India \quad
  \textsuperscript{3}Domyn, New York City, USA \\[3pt]
  \texttt{\{jacopo.biggiogera, nicolas.colombo, krati.saxena\}@domyn.com} \\
  \texttt{\{bhaskarjit.sarmah, stefano.pasquali\}@domyn.com}
}

\begin{document}
\maketitle

\begin{abstract}
High-stakes tabular pipelines rely on scattered, static quality checks that are hard to evolve and audit against policy intent. Regulated domains particularly suffer as the LLM-era automation raises expectations for traceability and accountability. We propose a vision for an \emph{ontology-grounded, repository-native} data quality layer in which rules are first-class artifacts: versioned, provenance-linked to governance texts, and re-playable through immutable execution traces. As a first step toward validating this vision, we present a concrete implementation of its agentic rule-discovery core with two rule acquisition modes (Full-Agentic and Guided) and a human-in-the-loop (HITL) review interface. We define an evaluation framework and run experiments addressing whether our fully autonomous, divide-and-conquer agentic rule discovery produces rules of the same quality and diversity as those specified by a human expert, reporting preliminary results on two real world industry datasets in the finance and retail domains.
\end{abstract}

\section{Introduction}

Tabular data drives high-stakes decisions, yet most organizations still rely on basic data quality (DQ) checks covering missingness, uniqueness, and simple range constraints.
Recent enforcement actions underscore the cost of undetected failures: a USD~\$100M settlement tied to AML compliance gaps~\citep{baker100m} and a USD~\$136M fine linked to data-management weaknesses~\citep{reutersciti136m}.
The hardest production failures arise from context-dependent constraints tied to domain semantics (limits that depend on market hours, settlement cycles, or cross-entity reconciliation) which are documented in governance policies but never fully operationalized, remaining handcrafted, scattered, and prone to drift as schemas evolve.

We propose a vision for an \emph{ontology-grounded, repository-native agentic DQ layer}: a system that automatically discovers rules from relational data, grounds them in semantic ontologies and governance artifacts, governs them through a human-in-the-loop lifecycle, and compiles approved rules into deterministic executable checks. Central to this vision is the idea that DQ rules should be treated as \emph{living rulebooks}: first-class, versioned artifacts with clear provenance, not scattered code buried in pipelines.

As a concrete step towards this vision, we have implemented the agentic rule-discovery core of the proposed system and designed an evaluation to answer the following research questions:

\begin{enumerate}
    \item Are rules discovered fully autonomously by the
          agentic system capable of approaching the quality ceiling set by those specified by a human?
    \item Are rules discovered fully autonomously by the
          agentic system as diverse and non-redundant as those
          specified by a human?
\end{enumerate}

\noindent In doing so, this paper makes three contributions:
\begin{itemize}
    \item \textbf{An agentic architecture for automatic rule discovery in relational database systems} .
    \item \textbf{An empirical evaluation of rule-discovery quality and diversity}.
    \item \textbf{A comprehensive vision for a full data-quality layer governing the complete rule lifecycle}.
\end{itemize}

\noindent We use two real industry derived datasets from the financial banking and retail fields to test the domain-agnostic design of the system.

\section{Related Work}
Our work draws on three threads: data agents, DQ tooling, and
ontology/audit infrastructure, but focuses on what they do not yet standardize: a shared, auditable representation of \emph{DQ rules themselves} and the automated discovery of such rules from data.

\textbf{Agentic data analytics.}
Recent data agents plan multi-step analyses and produce end-to-end outputs from heterogeneous data~\citep{sun2025agenticdata,lapin2025lightautods,qiao2025scaling}. Evaluation suites measure task success across analytics settings~\citep{nie2026dsgym,wang2025fdabench,padala2025tab}.
None treat DQ as a lifecycle problem in which rules are created, versioned, executed repeatedly, and updated under audit, nor do they evaluate the quality and diversity of automatically discovered constraints.

\textbf{Data-quality tooling.}
DQ stacks author and run checks and report violations~\citep{pancini2025engineering,machireddy2023data,patel2022automatic}. Learned approaches detect or repair errors and explain anomalies~\citep{dong2025automated,sattarov2022explaining}.
Formal constraint languages (denial constraints, SHACL/SHEx~\citep{thornton2019using})
offer expressive validators, but checks remain scattered in code with weak links to the policy text that motivated them and no replayable execution evidence for audit.
Crucially, none of these approaches address the \emph{automated discovery} of domain-grounded rules from relational data at scale.

\textbf{Ontologies, audit, and regulated-industry architectures.}
Auditable LLM pipelines argue for traceable, replayable decision-making~\citep{xia5775317evaluation,ojewale2026audit}.
Ontology-guided methods improve grounding and validation~\citep{guo2025ontology,chepurova2025wikontic,feng2025ontologyrag}.
Regulated-industry proposals motivate agentic DQ~\citep{tomar2024agentic,durgam2025scalable},
but do not define a concrete, testable contract linking ontologies, rules,
and execution traces, nor an evaluation of discovery quality.

We synthesize these threads by making rules and their evidence the central
auditable artifacts, introducing an agentic discovery architecture, and
providing an empirical evaluation of discovery quality and diversity.

\section{Problem Setting}

Organizations rely on tabular data (single tables, streaming logs, and multi-table warehouses) to power downstream reporting and high-stakes decisions. In finance, for example, small data errors can become material incidents: misreported exposures, missed AML signals, and model drift. End-to-end traceability is essential, requiring clear links from policy statements to implemented DQ checks.

\textbf{Current practice and limitations.}
Most production DQ remains a patchwork of simple checks spread across code, dashboards, and tools~\citep{saini2025unified}, with weak links to the semantic, temporal, and cross-table requirements in \autoref{tab:dq-requirements} and limited governance (versioning, provenance, auditability).
Beyond basic checks, policy-grounded semantic rules are costly and hard to identify and operationalize, lacking a principled automated mechanism to discover and formalize them.
These gaps motivate both our architecture and our research questions.

\begin{table}[t]
\small
\centering
\setlength{\tabcolsep}{4pt}
\begin{tabular}{p{0.30\columnwidth} p{0.65\columnwidth}}
\hline
\textbf{Requirement} & \textbf{What it enforces / example} \\
\hline
\textbf{Syntactic validity} & Types, ranges, formats across flat files and joins; referential integrity via keys/views. \\
\textbf{Semantic correctness} & Ontology-grounded invariants; e.g., notional $>0$ consistent with product type in finance via trade $\to$ instrument $\to$ client, or a medication dose within an age-adjusted safe range in healthcare. \\
\textbf{Temporal consistency} & No future-dated events relative to ingestion; consistent time-series snapshots across grains. \\
\textbf{Cross-table reconciliation} & Aggregates align with primaries; e.g., ledger totals vs. front-office trades, or risk vs. accounting views. \\
\hline
\end{tabular}
\caption{Core data-quality requirements for analytics-aware rules.}
\label{tab:dq-requirements}
\end{table}

\section{Implemented System}
\label{sec:architecture}

\subsection{Overview and design principles}

\begin{figure*}[t]
    \centering
    \includegraphics[width=0.95\textwidth]{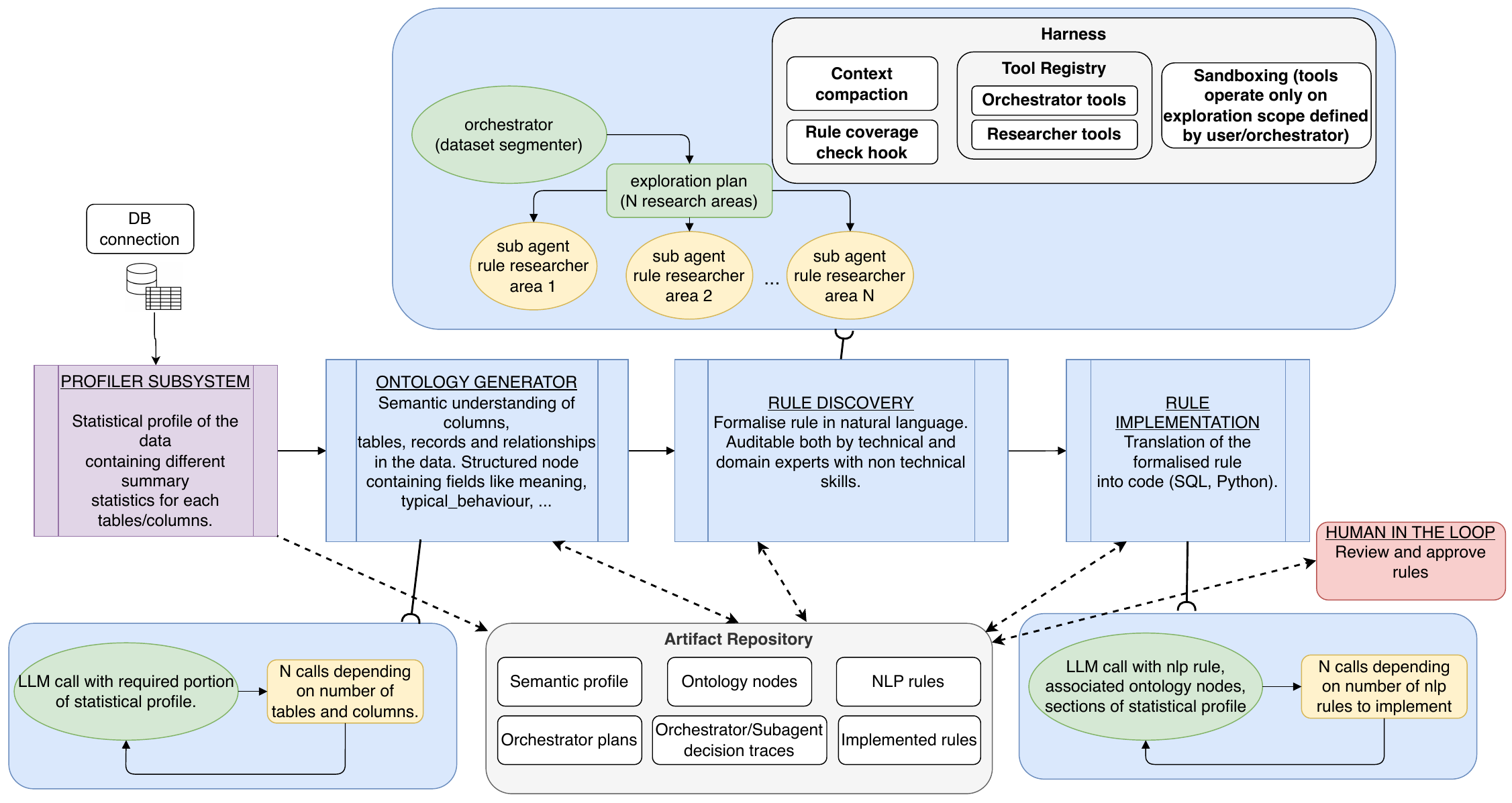}
    \caption{Implemented system overview.}
    \label{fig:architecture_implemented}
\end{figure*}

\autoref{fig:architecture_implemented} shows the four-stage pipeline we have implemented.
Data flows from a \textbf{Statistical Profiler} through an
\textbf{Ontology Generator} and a \textbf{Rule Discovery/Definition} module to a \textbf{Rule Implementer}.
Every stage reads from and writes to a central \textbf{Artifacts Repository}, which stores versioned semantic profiles, ontology nodes, natural language and implemented rules alongside orchestrator plans, subagent rule proposals, thoughts/tool calls made by every agent and recorded HITL decisions. A \texttt{LEARNINGS.MD} file is updated by a dedicated critic agent after each run, appending timestamped entries under sections for patterns to avoid, underexplored scopes, failure analyses, coverage notes, and heuristics that worked. This persistent memory informs subsequent rule-generation episodes. All coordination is repository-mediated: components write typed artifacts to the central Repository and read only from it, with no direct agent-to-agent message passing. This decoupling keeps lineage and audit straightforward and makes agents independently replaceable. Concrete artifact schema are given in Appendix due to page limits.

Three design principles govern the implementation.
\emph{Determinism before reasoning}: the profiler is LLM-free and produces the same output for the same input, providing a stable epistemic foundation.
\emph{Text-first rules}: rules are stored in natural language and reviewed before compilation, making them auditable also by domain experts without programming skills.
\emph{Repository-mediated coordination}: all component interaction is mediated through versioned artifacts, keeping lineage and audit straightforward.

\subsection{Statistical Profiler}

The profiler connects directly to the database and produces a
\emph{Statistical Evidence Pack}: a structured artifact containing distributional
summaries, null rates, cardinality estimates, temporal patterns, candidate
keys, and inter-column dependencies for every table and column in scope.
Being LLM-free, it is fully deterministic; downstream agents reason over
its outputs rather than raw rows, avoiding sampling bias and token-budget
issues on large tables.

\subsection{Ontology Generator}

For each table, entity, and column, a focused LLM call receives only the
relevant portion of the Statistical Evidence Pack.
A \emph{batched, asynchronous} strategy issues $N$ parallel calls, reducing wall-clock time proportionally to the number of objects.
Three harness properties are enforced: \emph{Context isolation}, each call
sees only its target object's statistics; \emph{Lifecycle hooks} run a duplicate-check function after each node
proposal; if the node already exists in the repository, the LLM call is
suppressed.
\emph{Evaluation} exposes proposed nodes to a human reviewer who can modify or delete them before they are committed.
Two usage modes are supported: \emph{automated generation}, which produces one node per table, entity, and column plus the three most salient vocabulary and within-table relationship nodes; and \emph{manual node} creation, where the user describes the desired node, selects a type and column association, and the agent either creates a new node or interprets the request as a modification of an existing one.

\subsection{Rule Discovery and Definition}
\label{sec:rga}
Rule discovery is the architectural core of the system and the focus of our evaluation.

\textbf{Full agentic mode.}
An \emph{orchestrator} receives the full profiled database and segments the column space into $N$ disjoint \emph{subspaces}, each assigned to a
\emph{researcher subagent}. Each subagent explores its scope autonomously and proposes rules until a \emph{coverage criterion} is met: every column in the subspace must be associated with at least one rule.
The orchestrator has access to lookup tools for table/column listings, profile summaries and ontology nodes. Finally, planning tools help it draft/modify a plan file to divide the search space in meaningful sub spaces and spawn workers; researcher subagents additionally have column-level profile queries, ontology lookups, a vector-DB search over business documents (governance grounding), and a rule-proposal tool that saves decisions to the artifact repository. During proposal each rule has the hard constraint of having to be linked to at least one ontology node in order for it to be considered valid.

The harness manages long-running execution through \emph{context compaction} (token-budget-triggered summarization with smart reinitialisation injecting uncovered columns and existing rules), \emph{coverage enforcement} (a
lifecycle hook terminates the loop when the criterion is met), and \emph{persistent state} (all outputs written to storage for run recovery from partial states).

\textbf{Guided mode.}
A domain expert describes the rules they already know should exist, optionally specifying the relevant columns and tables, and the system serializes the rule structure and generates code.
If no subset is specified, the orchestrator finds one subspace containing the relevant tables/columns and spawns a subagent to generate the described rule.
In the following experiments this mode serves as our human-expert baseline. The agent played strictly the role of \emph{serialization}: converting a human-identified constraint into the same schema and logic-block structure used in Full-Agentic mode, so that both conditions yielded structurally comparable artefacts. The rule mining act itself (exploring the dataset and identifying a candidate constraint together with the tables and columns it covers ) was performed entirely by the human actor.

\subsection{Rule Implementer and HITL Governance}

For each NLP rule selected for implementation, a single LLM call receives
the rule text, its associated ontology nodes, and the relevant portions of the statistical profile.
As with the ontology generator, a batched, asynchronous strategy is used and the system then translates each into executable SQL or Python code.
A HITL step allows the user to view the NLP rule and the generated code, modify the rule logic if needed, and re-trigger code generation.
Rules move through three lifecycle states: \emph{Proposed}, \emph{Approved}, \emph{Rejected}, with every transition recorded in the artifact repository.

\section{Evaluation}
\label{sec:evaluation}

A central capability for an agentic rule-discovery system is whether autonomous discovery can approach the quality and diversity ceiling set by human specified rules, thus making this the key first step for evaluation when developing said systems.
This is non-trivial: the full-agentic mode must be able to cover the entire column space and produce semantically varied rules. The divide-and-conquer strategy previously described in section \S\ref{sec:rga} introduces a specific risk: \emph{cross-agent leakage}: researcher subagents operating on adjacent subspaces may independently converge on semantically equivalent rules encoding the same constraint with different syntax. Our evaluation is designed to quantify both failure modes.

We compare \emph{Full Agentic} mode and \emph{Guided} mode, with the latter being a structural upper bound on both dimensions: quality (the human specifies intent directly) and diversity (user specifies rules manually one at a time to prevent redundancy and contradictions).

\subsection{Metrics}

\textbf{Good-rule rate.}
For dataset $D$ and mode $M$ $\in [A (automated), G (guided)]$ , let $N_M$ be the number of rules reviewed,
$A_M$ accepted as-is, $U_M$ accepted with modifications, and $R_M$ rejected.
We define:
\[
\hat{p}^M_{\mathrm{good}} = \frac{A_M + \alpha\, U_M}{N_M}, \quad \alpha = 0.5,
\]
where modified-but-approved rules receive partial credit through the $\alpha = 0.5$ weight.

\textbf{Information Divergence Score.}
Following from ~\citep{lee2025gemmas} we define an IDS measure adapted to our own domain to evaluate pairwise diversity across $N$ proposed rules:
\[
\mathrm{IDS} = 1 - \frac{1}{\binom{N}{2}} \sum_{i < j} \mathrm{sim}_{ij}
\]
where:
\[
\begin{aligned}
\mathrm{sim}_{ij} &= w_{\mathrm{syn}}\,\mathrm{Jac}(t_i, t_j)
+ w_{\mathrm{sem}}\,\cos(e_i, e_j) \\
&\quad + w_{\mathrm{scope}}\,\mathrm{Jac}(\mathit{s}_i, \mathit{s}_j)
\end{aligned}
\]

Pairwise similarity thus combines three components,
\begin{enumerate}
    \item \textbf{Syntactic}: token-level Jaccard similarity over rule logic and query text, $\mathrm{Jac}(t_i, t_j)$.
    \item \textbf{Semantic}: dense-embedding cosine similarity over rule logic and query text, $\cos(e_i, e_j)$.
    \item \textbf{Rule scope}: Jaccard similarity over the union of ontology nodes and table/column names referenced by each rule, $\mathrm{Jac}(s_i, s_j)$.
\end{enumerate}

IDS $\in [0,1]$ with a higher score signifying more diverse rules.

\subsection{Datasets}
We evaluate the system on two publicly available industry relational databases that capture real diversities in domain, schema and complexity.

\textbf{Berka (retail banking).}
The Berka dataset contains real, anonymized data from a Czech bank covering the period 1993--1998.
It comprises 8 tables (\texttt{account}, \texttt{client},
\texttt{disp}, \texttt{card}, \texttt{loan}, \texttt{order},
\texttt{trans}, and \texttt{district}) linked by foreign keys that model the full lifecycle of retail banking. The database contains 5,369 clients and over 1,000,000 transaction records, making it a rich, multi-table relational benchmark with realistic referential integrity constraints, coded
categorical fields, and cross-table financial consistency invariants. It is widely used in the relational data mining literature~\citep{li2023can, singh2025finch}.

\textbf{rel-HM (fashion retail).}
The rel-HM dataset contains real, anonymized data from retail purchases of HM products. It contains 3 tables (\texttt{customers}, \texttt{articles}, and \texttt{transactions}) with approximately 16.7 million purchases from 2018 to 2020. Despite its simpler schema, it presents high-cardinality categorical vocabularies, implicit referential constraints between articles and transactions, and cross-field consistency rules between product metadata fields \citep{robinson2024relbench}. As both dataset do not ship with any grounding documentation the rule discovery system ran without the governance-grounding search tool, using only the dataset statistics and the inferred ontology.

\subsection{Experimental protocol}
For each dataset, we ran both the Full Agentic mode and the Guided mode  with $k = 4$ reviewers, each yielding between 25-55 proposed rules per run and a total of 16 runs overall. Reviewers were mid/senior data engineering professionals with finance and retail domain experience.
Every reviewer authored one Guided run to which a Full agentic run was paired for rule reviewing load purposes. Rule assignment for review followed a blinded, mixed-mode, round-robin batch design: each rule was rated by exactly two reviewers other than its author, every batch mixed rules from both generation modes and reviewers were blind to which mode had produced each rule. We report the full block-construction protocol and validation checks in Appendix \S\ref{sec:review-protocol}. All rules were reviewed via the standard HITL flow and labeled accept/modify/reject. All the ontology and rule generation processes utilised Qwen3.5-397B-A17B as the underlying LLM model, chosen for its competitive performances, open source nature and generous context/output limits. A temperature of 0.5 was selected to allow for variability across exploration episodes.

\textbf{Experiment 1 (RQ1): Rule-generation quality by mode.}
We test for equivalence, $H_0$: $|\hat{p}^A_{\mathrm{good}} - \hat{p}^G_{\mathrm{good}}| \geq \Delta$ against
$H_1$: $|\hat{p}^A_{\mathrm{good}} - \hat{p}^G_{\mathrm{good}}| < \Delta$, by  estimating the shift in good-rule rate between modes,
$\hat{p}^{A}_{\mathrm{good}} - \hat{p}^{G}_{\mathrm{good}}$, per dataset and pooled across datasets. Because per-run scores are few and non-normal, we use a non-parametric two one-sided test (TOST): a Hodges--Lehmann (HL) shift estimator, exact confidence intervals obtained by inverting the Mann-Whitney statistic, rank-biserial effect sizes, and the TOST power achieved. The equivalence
margin was fixed before the analysis at the standard deviation of
Guided-mode run scores ($\Delta = 0.0556$), so that modes count as equivalent only if they differ by less than reviewers differ among themselves. We report the HL shift and its interval as our primary estimate, and the TOST verdict alongside its achieved power. Because every rule carries two independent ratings, we additionally quantify inter-rater agreement on the accept/modify/reject scale. We report raw agreement alongside two chance-corrected coefficients: ordinal Krippendorff's $\alpha$ and Gwet's AC2, both with $95\%$ bootstrap intervals. We report both because they behave differently under skewed marginals, and our verdict distribution is heavily skewed; relying on either alone would misstate the reliability of the labels.

\textbf{Matched-scope volume control.}
The two modes propose different rule volumes as the Full Agentic coverage criterion forces at least one rule per column generating roughly 1.6x the rules of the Guided mode (\autoref{tab:hitl-outcomes}). Good-rule rate is a precision-style measure, so comparing it across sets of very different size favors the mode that proposes fewer, safer rules. Since neither dataset ships with a reference rule set to compare against we repeat the equivalence test with a matched-scope volume control. We recorded the tables and columns referenced in each rule of all Guided mode runs and took the intersection across all four reviewers (a conservative proxy for the scope every reviewer agreed was rule-worthy). We then restricted the agentic rules set to that same scope. This removes the potentially low-value rules forced by the coverage criterion and makes volumes comparable (Agentic:Guided ratio $=$ 1:1 on Berka, 1:1.3 on H\&M). Scope matching is applied post hoc, at analysis time rather than at batch construction, so it does not affect what reviewers evaluated. We then recompute the RQ1 analysis on this matched set.

\textbf{Experiment 2 (RQ2): Diversity and partition effectiveness.}
We test for equivalence, $H_0$: $|\mathrm{IDS}_A - \mathrm{IDS}_G| \geq \Delta$ against
$H_1$: $|\mathrm{IDS}_A - \mathrm{IDS}_G| < \Delta$ by estimating the shift in Information Divergence Score between modes, $\mathrm{IDS}_A - \mathrm{IDS}_G$, using the same non-parametric TOST procedure, with the margin fixed a priori at the standard deviation of Guided mode IDS scores ($\Delta = 0.0202$). For each dataset and replicate we use pre-reviewed rules (to avoid human review masking the agentic run's inherent diversity through rule modification) and compute IDS per run. To ensure that any result is not an artefact of a single similarity component, we repeat the analysis across four weight configurations: equal weights $(w_{\mathrm{syn}}, w_{\mathrm{sem}}, w_{\mathrm{scope}}) = (1/3, 1/3, 1/3)$, semantic-heavy $(0.25, 0.5, 0.25)$, syntax-heavy $(0.5, 0.25, 0.25)$, and scope-heavy $(0.25, 0.25, 0.5)$.

\subsection{Results}
\label{sec:results}

\paragraph{RQ1: Rule-generation quality.}
\autoref{tab:rq1-summary} reports the mean good-rule rate per mode and dataset. The Full agentic mode achieves a higher mean good-rule rate on both datasets (Berka: 0.890 vs.\ 0.880; H\&M: 0.922 vs.\ 0.850), and the gap is larger on the rel-HM, the dataset with a simpler structure.

\begin{table}[t]
\small
\centering
\setlength{\tabcolsep}{4pt}
\begin{tabular}{llc}
\hline
\textbf{Dataset} & \textbf{Mode} & $\hat{p}_{\mathrm{good}}$ \\
\hline
\multirow{2}{*}{Berka}
  & Full agentic & 0.890 \\
  & Guided & 0.880 \\
\hline
\multirow{2}{*}{H\&M}
  & Full agentic & 0.922 \\
  & Guided & 0.850 \\
\hline
\multirow{2}{*}{Overall}
  & Full agentic & 0.906 \\
  & Guided & 0.865 \\
\hline
\end{tabular}
\caption{Mean good-rule rate across runs per dataset and mode.}
\label{tab:rq1-summary}
\end{table}

Full Agentic mode scored slightly higher than the Guided mode overall ($0.037$ Hodges--Lehmann shift, 90\% CI: $-0.008$ to $+0.089$), with a small shift for Berka ($+0.004$, 90\% CI: $-0.112$ to $+0.135$) and a larger one for H\&M ($+0.064$, 90\% CI: $+0.029$ to $+0.129$), where the interval excludes zero: on this dataset the four agentic runs separate completely from the four guided runs ($r = -1$). \autoref{tab:rq1-tost} reports the full analysis.
Equivalence is not established for any comparison since every CI extends beyond the set margin ($\Delta = 0.0556$). Achieved power is low throughout ($27.8\%$ overall), rendering the equivalence verdict uninformative
rather than simply unestablished, and thus we draw no conclusion of quality parity between the modes. The matched-scope volume control leaves the estimate essentially unchanged ($+0.037$, 90\% CI: $-0.027$ to $+0.094$), indicating that the volume asymmetry created by the coverage criterion is not what produces the agentic-favouring shift.

\begin{table}[t]
\footnotesize
\centering
\setlength{\tabcolsep}{2pt}
\begin{tabular}{lcccc}
\hline
\textbf{Comparison} & \textbf{HL} & \textbf{90\% CI} & $r$ & \textbf{Power} \\
\hline
Berka   & $+0.004$ & $[-0.112, +0.135]$ & $0$ & $<1\%$ \\
H\&M    & $+0.064$ & $[+0.029, +0.129]$ & $-1$ & $36.5\%$ \\
Overall & $+0.037$ & $[-0.008, +0.089]$ & $-0.50$ & $27.8\%$ \\
\hline
\multicolumn{5}{l}{\emph{Matched-scope control}} \\
Overall & $+0.037$ & $[-0.027, +0.094]$ & $-0.38$ & $3.5$ \\
\hline
\end{tabular}
\caption{Non-parametric TOST equivalence analysis for RQ1 good-rule rate (Full Agentic $-$ Guided); equivalence margin $\Delta = 0.0556$. HL: Hodges--Lehmann shift with exact CI from the inverted Mann-Whitney statistic; $r$: rank-biserial effect size (negative favor Full Agentic mode); Power: achieved TOST power. Runs required for 80\% power: 28 (Berka), 6 (H\&M), 16 (Overall). The matched-scope: rules with column scope shared by all Guided runs.}
\label{tab:rq1-tost}
\end{table}

\paragraph{Inter-rater agreement.}
\autoref{tab:irr} reports agreement over the $541$ doubly-rated rules.
Agreement is high: reviewers selected the same verdict on $77.3\%$ of rules, and Gwet's AC2 is $0.812$ overall, with consistent values across both datasets and both modes. Ordinal Krippendorff's $\alpha$ is lower ($0.174$) because $84.8\%$ of ratings are \texttt{accept}; when one category dominates, $\alpha$'s chance-correction term approaches the observed agreement and the coefficient is deflated. AC2 is a metric robust to exactly this situation, hence why both are reported to better understand the caveat around the ratings. Of note is that $\alpha$ is lower on agentic than on guided rules despite near-identical raw agreement ($76.8\%$ vs. $78.1\%$). The gap is largely arithmetic: accept is even more dominant on agentic rules ($86.0\%$ vs. $82.6\%$, \autoref{tab:hitl-outcomes}), which raises the agreement expected by chance alone and leaves less credit for $\alpha$ to assign. Appendix~\S\ref{sec:irr-detail} gives the full breakdown.

\begin{table}[t]
\footnotesize
\centering
\setlength{\tabcolsep}{2pt}
\begin{tabular}{lcccc}
\hline
\textbf{Subset} & $n$ & \textbf{Exact} & $\alpha$ \textbf{[95\% CI]} & \textbf{AC2 [95\% CI]} \\
\hline
Overall & 541 & 77.3 & 0.174 [.074,.279] & 0.812 [.770,.851] \\
\hline
Berka   & 317 & 76.0 & 0.165 [.035,.292] & 0.810 [.754,.858] \\
H\&M    & 224 & 79.0 & 0.189 [.021,.350] & 0.815 [.747,.872] \\
\hline
Agentic & 340 & 76.8 & 0.061 [$-$.056,.187] & 0.816 [.764,.864] \\
Guided  & 201 & 78.1 & 0.326 [.150,.484] & 0.805 [.730,.868] \\
\hline
\end{tabular}
\caption{Inter-rater agreement over doubly-rated rules. \textbf{Exact}: \% rules where both reviewers gave same verdict.
$\alpha$: ordinal Krippendorff's $\alpha$; AC2: Gwet's AC2; both with $95\%$
bootstrap intervals. $\alpha$ is deflated relative to AC2 because
\texttt{accept} accounts for $84.8\%$ of all ratings.}
\label{tab:irr}
\end{table}

\paragraph{RQ2: Rule diversity and partition effectiveness.}
\autoref{tab:rq2-ids} reports mean IDS per mode across four weight
configurations. IDS values are nearly identical between modes: under
semantic-heavy weights both achieve a mean IDS of 0.648, and the largest
absolute difference across configurations is 0.010 (scope-heavy, favouring the agentic mode). The sign of $\Delta$ changes across weight
configurations, indicating no consistent direction of difference.

\begin{table}[t]
\small
\centering
\setlength{\tabcolsep}{4pt}
\begin{tabular}{lccc}
\hline
\textbf{Weight config} & $\mathrm{IDS}_A$
  & $\mathrm{IDS}_G$ \\
\hline
Equal             & 0.732 & 0.729 \\
Sem.-heavy    & 0.648 & 0.648 \\
Syn.-heavy      & 0.749 & 0.750 \\
Scope-heavy       & 0.798 & 0.788 \\
\hline
\end{tabular}
\caption{Mean IDS across datasets ($n_A = n_G = 8$ runs per mode).}
\label{tab:rq2-ids}
\end{table}

\autoref{tab:rq2-tost} results support equivalence under every weight
configuration, with small effect sizes ($|r| \le 0.28$), moderate power (44--71\%) and confidence intervals falling entirely within the a priori margin ($\Delta = 0.0202$). The result holds across all four configurations, highlighting how it isn't an artefact of any single similarity component of IDS. This provides a result in support of RQ2 and thus the system's capability to avoid redundancy while retaining exploratory parallelization capabilities of the complex information spaces that relational databases represent. Column space division across concurrently operating subagents doesn't measurably reduce semantic diversity of the resulting rule set relative to human specification.

\begin{table}[t]
\footnotesize
\centering
\setlength{\tabcolsep}{2pt}
\begin{tabular}{lcccc}
\hline
\textbf{Weights} & \textbf{HL} & \textbf{90\% CI} & $r$ & \textbf{Power} \\
\hline
Equal    & $-0.003$ & $[-0.010, +0.012]$ & $+0.16$ & $57\%$ \\
Semantic & $-0.005$ & $[-0.015, +0.008]$ & $+0.22$ & $44\%$ \\
Syntax   & $-0.005$ & $[-0.016, +0.010]$ & $+0.25$ & $71\%$ \\
Scope    & $+0.003$ & $[-0.004, +0.019]$ & $-0.28$ & $53\%$ \\
\hline
\end{tabular}
\caption{Non-parametric TOST equivalence analysis for RQ2 IDS
(Full Agentic $-$ Guided); equivalence margin $\Delta = 0.0202$. HL: Hodges--Lehmann shift with exact CI from
the inverted Mann-Whitney statistic; $r$: rank-biserial effect size. Runs required for 80\% power: 11 (Equal), 13 (Semantic), 9 (Syntax), 11 (Scope).}
\label{tab:rq2-tost}
\end{table}
\section{Vision: Closed-Loop DQ Layer}
\label{sec:vision}

The prototype implemented addresses rule discovery and governance. The broader vision extends this into a full closed-loop DQ layer with two additional components (full details and system diagram in Appendix \S\ref{sec:appendix-EDRA-spec}, \S\ref{sec:appendix-rcada-spec} and \autoref{fig:architecture}).

\textbf{Execution Dashboard and Reporting Agent (EDRA)} runs checks idempotently on stable dataset snapshots, and writes immutable \emph{ExecutionTraces} with violation counts, sample failing rows, runtime, and provenance (rule version, ontology snapshot, dataset snapshot). It then aggregates ExecutionTraces into interactive health dashboards, severity-weighted dataset scores, rule-stability trend views, and audit-ready exports, providing a single pane of glass for data owners, risk, compliance, and engineering.

\textbf{Root Cause Analysis and Detective Agent (RCADA)} reads
ExecutionTraces, ontology risk metadata, dataset lineage, and profiler
trends to produce ranked root-cause hypotheses using an explicit cause
taxonomy (provider outage, pipeline regression, upstream drift, contextual
event, ontology mismatch, rule mis-specification), bounded near-term failure forecasts, and non-binding prescriptive suggestions for steward review.

Together, these components close the feedback loop: EDRA executes and summarizes, RCADA diagnoses and proposes, humans decide, and the repository records updates for the next cycle.
Because all rules and traces are linked to stable ontology node IDs, the system remains robust to schema and provider changes by reasoning over semantic entities rather than raw column names.


\section{Discussion}
We have proposed a vision for an ontology-grounded, versionable DQ system, taking a concrete first step by implementing and evaluating its agentic rule-discovery core and the HITL components needed for auditability/reproducibility requirements in regulated domains.
Our results on two real-world databases show that the full-agentic mode produces rules of comparable diversity to the human baseline, supporting the semantic-coverage premise of the vision. 
However, the lack of established equivalence combined with the underpowered nature of the test for the first research question leaves it open, and we therefore make no current conclusions towards the quality parity between modes which will be further explored in future works.
The diversity finding is consistent across two datasets from different domains (financial/retail), which is encouraging but preliminary evidence of domain-agnostic transferability.

\textbf{Limitations and future steps}
We identify five limitations that qualify these results.
First, rule \emph{complexity} is not yet measured: our metrics capture quality and diversity but not whether automatically discovered rules are systematically simpler than human-specified ones, which would represent a meaningful gap in expressive power. Future evaluations will focus on quantifying also this dimension building on top of IDS and carrying out more comprehensive analysis on the metric's correlation to human judgment.
Second, both datasets might share characteristics across domains (ex both have a transactional nature) that have not been controlled for and which may not highlight the true limits of domain transferability; broader domain coverage is needed to test the system's limits. Expanding to datasets across existing relational benchmarks and synthetic data generation benchmarks is a priority for future work, with the longer-term goal of curating a benchmark pack including data snapshots, evolution scripts, governance text, and gold rule catalogs to share with the broader research community to support re-playable, lifecycle-aware evaluation specifically of Agentic DQ systems.
Third, scaling reviewers is inherently expensive, while small groups introduce issues with statistical power, as seen for RQ1 where a low power does not make it possible to draw conclusions on equivalence, or for RQ2 where a higher power would allow to draw even more confident conclusions. Future work will thus complement a larger reviewer set with automated evaluation based on rule execution outcomes and expert-validated reference gold rule sets. Fourth, compute budgets for these tests restricted us to a single LLM and temperature setting. We recognize the need for future model/temperature ablation test to better understand the robustness and transferability of the proposed harness. 
Finally, the current evaluation covers only the rule-discovery component; the full vision (execution+dashboarding, and root-cause analysis) remains to be implemented and evaluated end-to-end, thus at present we do not claim any empirical demonstration of the full architecture that we leave for future work.
\newline
\textbf{Open problems.}
Grounding noisy governance text into calibrated ontology updates, mapping abstract concepts across heterogeneous schemas, handling schema evolution without destabilizing linked rules, and measuring false positives at scale remain key open challenges we hope this framework can help start to address
systematically.
\section{Acknowledgments}
The authors would like to thank Aniket Joshi for having sourced the databases and prepared them for use, while Michael Santarossa and Roberto Varinelli for their feedback and precious advice on the work carried out for this paper.


\bibliographystyle{acl_natbib}
\bibliography{custom}

\appendix
\clearpage
\section*{Appendix}

\section{Beyond Traditional Data-Quality Checks}
\label{sec:appendix-traditional-vs-semantic}
Most production pipelines start with generic checks that do not depend on the domain. Common examples of traditional data quality checks are completeness and missingness, uniqueness and duplicate detection, data type and parsing validity, format checks such as date and identifier patterns, value ranges, basic outliers, and referential integrity across keys. These checks are necessary, but they often miss errors that only become visible when the data is interpreted in its real business meaning.

We refer to the domain-specific sophisticated data quality checks as \emph{policy grounded semantic rules}. These are domain dependent rules that use real world constraints and governance text such as policies, data dictionaries, and reporting definitions. They are often conditional, time aware, and cross table. They also need provenance so reviewers can see which document or evidence justified the rule. Some examples are shown in \autoref{tab:semantic-rules}.

\begin{table}[h]
\small
\centering
\setlength{\tabcolsep}{3pt}
\begin{tabular}{p{0.26\columnwidth} p{0.50\columnwidth} p{0.18\columnwidth}}
\hline
\textbf{Rule type} & \textbf{Finance example} & \textbf{Grounding} \\
\hline
Calendar and market hours & No equity trades on Sunday. Trades outside market hours must be flagged or explained. & Market calendar, policy \\
Temporal consistency & Settlement date must be after trade date. For many products it follows a fixed lag such as T plus 2, with exceptions. & Policy, product spec \\
Cross field meaning & Notional should match quantity times price within tolerance and currency scaling. & Domain math, evidence \\
Entity state & Counterparty must be active on trade date. Trades for blocked accounts must be rejected. & Reference data, policy \\
Cross table reconciliation & Trade totals must reconcile with positions and ledger aggregates at the portfolio level. & Reporting definition \\
\hline
\end{tabular}
\caption{Examples of policy grounded semantic rules that go beyond traditional checks.}
\label{tab:semantic-rules}
\end{table}

In addition, we also frame our data quality system around the four classic analytics types: descriptive, diagnostic, predictive, prescriptive that are a standard in the field.
\autoref{tab:agent-analytics} maps the full envisioned agents stack to said analytics types.

\begin{table*}[t]
\small
\centering
\setlength{\tabcolsep}{2.5pt}
\begin{tabular}{p{0.15\textwidth} p{0.20\textwidth} p{0.20\textwidth}
                p{0.20\textwidth} p{0.20\textwidth}}
\hline
\textbf{Agent} & \textbf{Descriptive} & \textbf{Diagnostic}
               & \textbf{Predictive} & \textbf{Prescriptive} \\
\hline
Ontology Gen. &
  Grounds columns into entities, measures, and units. &
  Adds semantic context for anomaly explanation. &
  Marks forecastable measures and derived metrics. &
  Enables domain-aware imputation and unit normalization. \\
Rule Discovery &
  Drafts checks with clear summaries and coverage metrics. &
  Encodes conditional/composite rules for diagnosis. &
  Proposes drift-sensitive and stability-aware rules. &
  Suggests adaptive thresholds and remediation-aware rules. \\
Implementer &
  Translates approved rules into executable code. &
  Generates diagnostic checks capturing failure evidence. &
  Produces code artifacts that accumulate execution history for forecasting. &
  Emits validated, controlled fixes to apply safeguards/remediation. \\
Human review &
  Confirms expected vs.\ concerning signals. &
  Confirms or rejects proposed causes. &
  Decides which predicted risks are actionable. &
  Approves, edits, or blocks remediation strategies. \\
EDRA &
  Executes checks and visualizes quality metrics and trends. &
  Captures failure evidence and supports drill-down/lineage views. &
  Collects execution history and displays risk scores and drift forecasts. &
  Presents recommended actions for review. \\
RCADA &
  Consumes descriptive trace summaries. &
  Produces ranked RCA hypotheses with evidence. &
  Forecasts future failures from recurring patterns. &
  Suggests targeted interventions and rule updates. \\
\hline
\end{tabular}
\caption{How agents and human review support descriptive, diagnostic,
predictive, and prescriptive behavior.}
\label{tab:agent-analytics}
\end{table*}

\section{Formal Artifact Specifications}
\label{sec:appendix-dsl}

\subsection{Ontology specifications}
\label{sec:appendix-ontology-spec}
We propose a clear, minimal, and extensible ontology specification for the ontology-enhanced agentic data quality system. The ontology is designed to be LLM-friendly, human-readable, and agent-operable, while remaining robust across domains. Design goals include stability (a fixed structural contract so agents can interoperate), expressiveness (rich semantics captured primarily in natural language), partial observability (nodes may be incomplete or uncertain by design), domain generality (the same ontology spine works across tabular domains), and LLM-first reasoning (semantics optimized for textual reasoning).

The ontology is a typed semantic graph: each node represents a semantic unit in tabular data, and relationships between nodes are captured explicitly or via references. The ontology supports exactly five node types, and this set is closed by design; new semantic distinctions should be expressed via \texttt{semantic\_role}, not by adding node types (\autoref{tab:ontology-node-types}).
\begin{table}[t]
\small
\centering
\setlength{\tabcolsep}{4pt}
\begin{tabular}{p{0.22\columnwidth} p{0.72\columnwidth}}
\hline
\textbf{Node type} & \textbf{Purpose} \\
\hline
Dataset & Represents a table, file, or view as a whole. \\
Entity & Represents a real-world object described by rows. \\
Attribute & Represents a column or derived field. \\
Relationship & Represents row-to-row or table-to-table linkage. \\
Vocabulary & Represents a value space or code system. \\
\hline
\end{tabular}
\caption{Exhaustive ontology node types.}
\label{tab:ontology-node-types}
\end{table}

Every ontology node must conform to the minimal schema below.
\begin{quote}
\small
\begin{verbatim}
{
  "node_id": "string",
  "node_type": "Dataset | Entity | Attribute |
   Relationship | Vocabulary",
  "name": "string",
  "semantic_role": "string",
  "confidence": "float",
  "sources": ["data | domain | human 
  | external"],
  "version": "string",
  "meaning": {
    "description": "string"
  }
}
\end{verbatim}
\end{quote}

Field meanings are summarized in \autoref{tab:ontology-node-fields}.
\begin{table}[t]
\small
\centering
\setlength{\tabcolsep}{4pt}
\begin{tabular}{p{0.24\columnwidth} p{0.70\columnwidth}}
\hline
\textbf{Field} & \textbf{Meaning} \\
\hline
\texttt{node\_id} & Stable unique identifier. \\
\texttt{node\_type} & One of the five exhaustive types. \\
\texttt{name} & Human-readable label. \\
\texttt{semantic\_role} & Open-ended semantic classification. \\
\texttt{confidence} & Belief score in correctness, from 0 to 1. \\
\texttt{sources} & Provenance of semantic evidence. \\
\texttt{version} & Ontology version tag. \\
\texttt{meaning.\allowbreak description} & Canonical semantic explanation. \\
\hline
\end{tabular}
\caption{Field semantics for the minimal ontology node schema.}
\label{tab:ontology-node-fields}
\end{table}

Beyond \texttt{meaning.description}, we support a soft semantic envelope for natural-language keys such as \texttt{constraints}, \texttt{typical\_behavior}, \texttt{edge\_cases}, and \texttt{risk\_notes}. For example:
\begin{quote}
\small
\begin{tabular}{p{0.95\linewidth}}
\ttfamily
"meaning": \{\\
\ \ "description": "Executed equity trades table.",\\
\ \ "constraints": "Each row is one executed trade.",\\
\ \ "typical\_behavior": "High volume during market hours.",\\
\ \ "edge\_cases": "After-hours trades and corrections.",\\
\ \ "risk\_notes": "Used for risk and regulatory reporting."\\
\}\
\end{tabular}
\end{quote}
Agents may add new keys such as \texttt{regulatory\_context} or \texttt{temporal\_behavior}, but must not remove or overload existing keys.

\subsection{Rules specifications}
\label{sec:appendix-artifact-repo-spec}
A structured, versioned rule repository works in tandem with the ontology. It is where data quality knowledge is proposed, reviewed, executed, audited, and evolved.
It serves as a staging ground for agent-generated rules, a human control point for approval and refinement, a persistent memory of data quality intent, a bridge from ontology to execution, and an audit log for governance. We propose core principles of being ontology-linked, text-first, lifecycle-aware, provenance-rich, and executable without being bound to any one runtime. Every rule must conform to a minimal schema.

\begin{quote}
\small
\begin{tabular}{p{0.95\linewidth}}
\ttfamily
\{\\
\ \ "rule\_id": "string",\\
\ \ "rule\_name": "string",\\
\ \ "linked\_nodes": ["ontology\_node\_id"],\\
\ \ "rule\_type": "descriptive | diagnostic | predictive | prescriptive",\\
\ \ "status": "draft | approved | deprecated",\\
\ \ "confidence": "float",\\
\ \ "generated\_by": "agent | human",\\
\ \ "version": "string",\\
\ \ "logic": \{\\
\ \ \ "intent": "string"\\
\ \ \}\\
\}\
\end{tabular}
\end{quote}

The fields carry the following meanings: \texttt{linked\_nodes} ties the rule to one or more ontology nodes; \texttt{rule\_type} captures the analytics orientation; \texttt{status} captures lifecycle state; \texttt{confidence} captures belief in correctness; and \texttt{generated\_by} records origin.

Rule logic is stored as typed natural-language blocks.
\begin{quote}
\small
\begin{tabular}{p{0.95\linewidth}}
\ttfamily
"logic": \{\\
\ \ "intent": "What quality property this rule enforces",\\
\ \ "condition": "When the rule applies",\\
\ \ "expectation": "What must hold true",\\
\ \ "severity": "Impact of violation",\\
\ \ "remediation": "Suggested corrective action",\\
\ \ "rationale": "Why this rule exists"\\
\}\
\end{tabular}
\end{quote}
Agents may add blocks such as \texttt{regulatory\_basis}, \texttt{statistical\_basis}, or \texttt{exceptions}, but must not redefine existing keys. Rules move through explicit states (Proposed, Approved, Rejected). State transitions are recorded as provenance events. Every rule must reference at least one ontology node; rules inherit meaning from linked nodes, and ontology updates may trigger rule re-evaluation. Example: \texttt{"linked\_nodes": ["col:trade\_amount", "entity:trade"]}.

\subsection{Profiler subsystem specifications}
\label{sec:appendix-profiler-spec}
The Profiler Subsystem is a deterministic, non-LLM component that extracts complete, coverage-aware evidence summaries from raw tabular data. It is the epistemic foundation of the system. Agents reason over profiler outputs rather than raw rows, which can be too large to route through an agent loop. The intended division of labor is that profilers compute facts, and agents interpret meaning.

Profiler inputs are raw tables or files, file-format metadata, execution configuration, and optional calendars or controlled business context. The profiler emits a \emph{Statistical Evidence Pack}, a structured artifact saved to the Artifact repository and consumed as immutable evidence by Ontology generation. We summarize the main profiling modules in \autoref{tab:profiler-modules}. The Statistical Evidence Pack is machine-readable; examples are shown below.

\begin{table}[t]
\small
\centering
\setlength{\tabcolsep}{3pt}
\begin{tabular}{p{0.22\columnwidth} p{0.70\columnwidth}}
\hline
\textbf{Module} & \textbf{What it outputs} \\
\hline
Structural & Schema facts such as column names, physical types, counts, declared keys, partitions. \\
Distributional & Global numeric and categorical summaries from full scan or distributed aggregation. \\
Temporal & Frequency, gaps, weekday or holiday patterns, rolling rates with confidence. \\
Relationship & Dependencies, candidate foreign keys, correlations, ordering constraints. \\
Quality signals & Null rates, duplicates, declared constraint counts, drift indicators. \\
\hline
\end{tabular}
\caption{Profiler modules and their evidence outputs.}
\label{tab:profiler-modules}
\end{table}

\begin{quote}
\small
\begin{tabular}{p{0.95\linewidth}}
\ttfamily
\{\\
\ \ "row\_count": 125000000,\\
\ \ "columns": \{\\
\ \ \ "trade\_id": "string",\\
\ \ \ "trade\_date": "date",\\
\ \ \ "trade\_amount": "decimal"\\
\ \ \}\\
\}\
\end{tabular}
\end{quote}
\begin{quote}
\small
\begin{tabular}{p{0.95\linewidth}}
\ttfamily
\{\\
\ \ "trade\_amount": \{\\
\ \ \ "p50": 120000,\\
\ \ \ "p99": 5000000,\\
\ \ \ "max": 87000000,\\
\ \ \ "negative\_rate": 0.0\\
\ \ \}\\
\}\
\end{tabular}
\end{quote}
\begin{quote}
\small
\begin{tabular}{p{0.95\linewidth}}
\ttfamily
\{\\
\ \ "trade\_date": \{\\
\ \ \ "inferred\_frequency": "daily",\\
\ \ \ "weekday\_gaps": ["Sunday"],\\
\ \ \ "max\_gap\_days": 2,\\
\ \ \ "calendar\_alignment": "NYSE",\\
\ \ \ "confidence": 0.97\\
\ \ \}\\
\}\
\end{tabular}
\end{quote}

The profiler records coverage and sampling policy explicitly.
\begin{quote}
\small
\begin{tabular}{p{0.95\linewidth}}
\ttfamily
\{ "profiling\_mode": "full\_scan", "coverage": "100\%", "sampling": "none" \}\
\end{tabular}
\end{quote}

If sampling is required, the policy and limitations are recorded. Profiler outputs are deterministic for a given dataset version. Profiling code versions are logged. The Evidence Pack is saved to the Artifacts Repository and reused by downstream agents. Profilers never consume ontology or rules.

\subsection{Ontology generation specifications}
\label{sec:appendix-sudoa-spec}
Ontology generation transforms profiler evidence into a semantic understanding of the dataset and emits ontology creation or update operations. It never reasons over raw rows at scale and instead consumes coverage-aware evidence summaries from the Statistical Evidence Pack.
Ontology generation reads the Statistical Evidence Pack from the Artifacts Repository. Mandatory inputs include structural metadata, distributional profiles, temporal profiles, and relationship and dependency profiles. Optional inputs include user hints, and controlled documentation. A key design principle is that Ontology generation never ingests raw rows; it reasons only over deterministic profiler evidence to avoid sampling bias and preserve temporal semantics.

The system emits a set of Ontology Operations as structured outputs to force the LLM to conform to the node schema (transactional \texttt{CREATE\_NODE} and \texttt{UPDATE\_NODE} operations that update the ontology namespace). For example, from a column profile for \texttt{trade\_amount}, the system can infer a monetary measure role and emit a manifest entry plus a node operation.
\begin{quote}
\small
\begin{tabular}{p{0.95\linewidth}}
\ttfamily
\{\\
\ \ "column": "trade\_amount",\\
\ \ "semantic\_role": "monetary\_measure",\\
\ \ "meaning": \{\\
\ \ \ "description": "Executed notional value of a trade.",\\
\ \ \ "constraints": "Must be non-negative.",\\
\ \ \ "risk\_notes": "Errors impact exposure and reporting."\\
\ \ \},\\
\ \ "confidence": 0.92\\
\}\
\end{tabular}
\end{quote}

Ontology operations are selected based on existing ontology match confidence. A new ontology namespace is created if no match is strong enough or isolation is requested. Otherwise Ontology generation updates existing nodes when new evidence refines meaning, constraints, units, or relationships.

\begin{figure*}[t]
    \centering
    \includegraphics[width=0.95\textwidth]{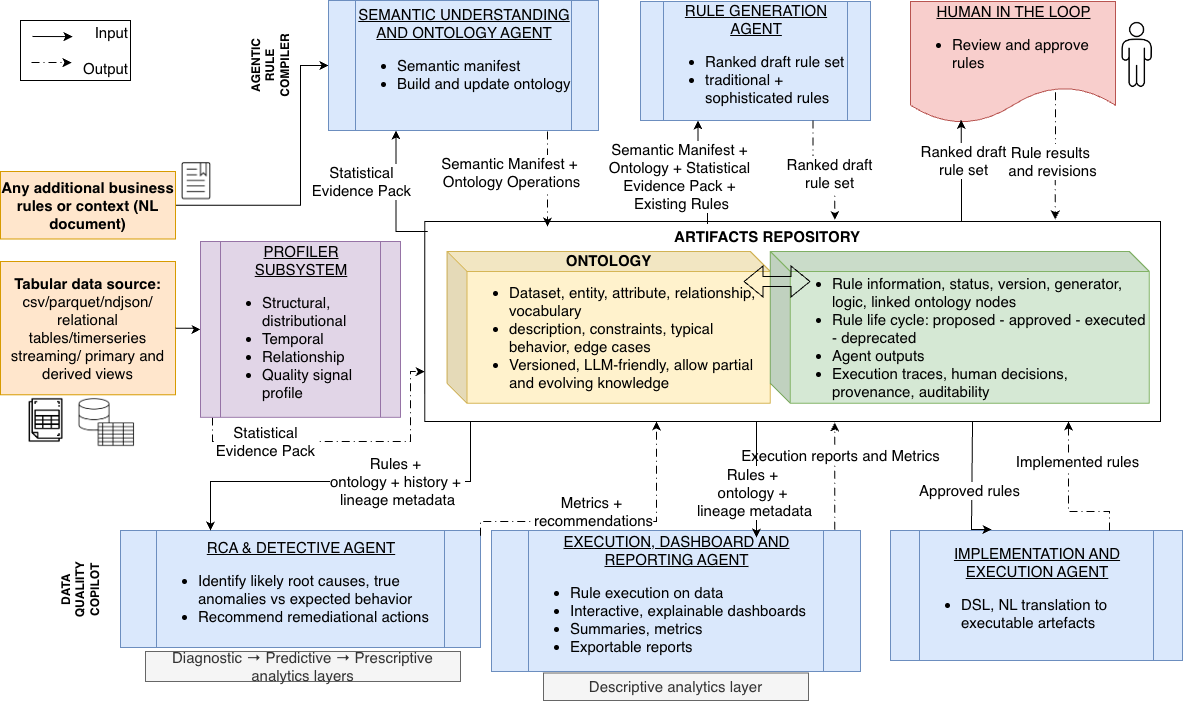}
    \caption{Full envisioned system overview: all components sharing
    a central Artifacts Repository across the full DQ lifecycle.}
    \label{fig:architecture}
\end{figure*}

\subsection{EDRA specifications (Vision)}
\label{sec:appendix-EDRA-spec}
EDRA is the interpretation and decision interface of the system. It is read-heavy, analytics-heavy, and explanation-focused. EDRA does not create rules and does not execute checks. Instead, it aggregates artifacts and human review in the Artifacts Repository, runs checks idempotently on stable dataset snapshots, writes immutable \emph{ExecutionTraces} with violation counts, sample failing rows, runtime, and provenance (rule version, ontology snapshot, dataset snapshot), computes higher-level analytics, and presents actionable, auditable views to stakeholders (data owners, risk, compliance, engineering).

EDRA's primary goal is to provide a single pane of glass for data-quality health, make failures explainable and traceable, support governance and audit, enable trend analysis and prioritization, and close the loop from rule failure to decision and remediation.

EDRA consumes read-only artifacts from the Artifacts Repository: rule definitions and metadata, ExecutionTraces and execution timelines, ontology references (node IDs, meanings, risk notes), and human decisions (approve, reject, modify with comments). Optional lineage metadata can be used to surface source-to-transformation context in investigations.

EDRA produces interactive dashboards, exportable audit-ready reports (e.g., JSON or CSV), alerts and summaries, dataset-level risk scores and trend forecasts, and non-binding prescriptive suggestions such as remediation playbooks or rule-tuning proposals. It organizes insights explicitly as descriptive (what happened), diagnostic (why), predictive (what is likely next), and prescriptive (what to do) analytics over the data-quality system itself.

A minimal EDRA UI includes (i) an Overview Dashboard with dataset health, top failing rules, and affected ontology nodes; (ii) a Rule Catalog with filters by dataset, ontology node, severity, status, and last execution; (iii) a Failure Explorer that shows the rule definition and rationale, violation statistics and samples, linked ontology meaning and risks, and supporting context such as lineage when available; (iv) Temporal and trend views for failure rate, recurrence, and stability; and (v) an Audit and governance panel that replays the rule lifecycle from draft to approved, modified, and deprecated with evidence exports.

EDRA computes severity-weighted dataset health, failure rate and recurrence, rule stability over time, coverage (the fraction of rows evaluated by active rules), and ontology risk exposure by aggregating the risk weights of affected ontology nodes.

\subsection{RCADA specifications (Vision)}
\label{sec:appendix-rcada-spec}
RCADA closes the loop. Given rule failures and execution traces, it explains why violations happened and what to do next by combining historical repository context, profiler trends, lineage, and optional external signals. RCADA primarily operates in the diagnostic, predictive, and prescriptive analytics layers; descriptive summaries are surfaced by EDRA.

For a set of failing rules and datasets, RCADA (i) identifies and ranks likely root causes, (ii) distinguishes true anomalies from contextual or expected behavior, and (iii) recommends concrete, human-controlled remediation actions with calibrated confidence.

RCADA reads (a) ExecutionTraces (failure rates, example rows, timestamps, runtime), (b) rule metadata (severity, linked ontology nodes, prior stability), (c) ontology metadata (semantic meaning, risk notes, provider dependencies), (d) artifact history (past outcomes and human decisions on similar failures), (e) dataset lineage (provider, pipeline stage, schema versions, deploy tags), and (f) profiler time series (null rates, distributions, cardinality trends). Optional external context includes calendars/holidays, outage logs, and governance or regulatory change notes.

RCADA writes a structured RCA Report artifact (root-cause candidates with evidence, classification such as contextual anomaly vs. likely incident, recommended actions with owners and urgency, and confidence/uncertainty). It can also emit bounded forecasts of near-term recurrence risk and non-binding prescriptive suggestions (e.g., rule refinement proposals or temporary usage restrictions) for human review.

RCADA uses an explicit and explainable cause taxonomy (e.g., provider outage, pipeline regression, upstream drift, contextual event, ontology mismatch, rule mis-specification). Candidates are scored using temporal alignment, cross-source agreement when comparable signals exist, historical recurrence, and ontology risk alignment. RCADA also uses human decisions as weak labels to estimate likely false positives and to flag systematically over-strict rules.

Example: Rule \texttt{R-042} (\texttt{trade\_amount} non-negative) shows a failure spike after \texttt{2026-01-25}. RCADA attributes 97\% of violations to \texttt{source=OTC\_FEED\_X}, notes ontology risk for PnL and regulatory reporting, and recommends an engineering escalation plus a steward-reviewed source-conditional exception with confidence.


\section{Blinded Rule Review Protocol}
\label{sec:review-protocol}

This appendix specifies the blinded, mixed-mode, overlapping-assignment
review design referenced in \S\ref{sec:evaluation}, and the reviewer pool
that carried it out. The design was built to satisfy three requirements: every rule is rated by exactly two different reviewers (enabling inter-rater agreement to be reported), no
reviewer ever rates a rule they authored and both generation modes are present
in every batch (so mode is never confounded with batch or rater identity).

\subsection{Block design}
Rules from each dataset were split into four blocks (A--D). Each block
contains the rules authored by exactly two of the four reviewers and is
rated by the two reviewers whose rules it does \emph{not} contain:

\begin{center}
\small
\begin{tabular}{ccc}
\hline
\textbf{Block} & \textbf{Authored by} & \textbf{Rated by} \\
\hline
A & R2, R3 & R1, R4 \\
B & R3, R4 & R1, R2 \\
C & R4, R1 & R2, R3 \\
D & R1, R2 & R3, R4 \\
\hline
\end{tabular}
\end{center}

Critically, both modes produced by a given author travel into the same
block together: Block A, for instance, contains both R2's and R3's Full
Agentic \emph{and} Guided rules. An author's agentic output is never routed
to a different block than their guided output, since splitting them would
let a between-author skill difference masquerade as a between-mode effect.
Within each author's allotment, rules are sorted by mode and dealt
alternately across the author's two permitted blocks, so that each mode is
split evenly rather than concentrated in one block. Presentation order
within a batch is shuffled independently per (block, reviewer) pair, and the
resulting position of each rule is logged.

This yields, per dataset, a connected co-rating cycle across reviewers, (R1,R4), (R1,R2), (R2,R3), (R3,R4), equal review workload across
reviewers, and every rule rated by exactly two people who did not write it.
One limitation follows directly from having four reviewers and two raters
per block: each run is seen by three of the four reviewers rather than all
four, and the pairs (R1,R3) and (R2,R4) never co-rate directly. Krippendorff's
$\alpha$ remains valid over the connected graph, but we do not report a
complete $4 \times 4$ pairwise agreement matrix, since two of the six pairs
were never jointly observed.

\subsection{Validation}
Before release, every batch was checked against five automated assertions:
exhaustive, non-overlapping coverage of all rules across blocks; exactly two
ratings per rule; no reviewer assigned to a block containing their own
authored rules; both modes present in every block; and both modes present
for every author within every block they occupy.

\subsection{Inter-rater agreement}
\label{sec:irr-detail}
The design yields $541$ doubly-rated rules and $1{,}082$ ratings.
\autoref{tab:irr} reports the coefficients; this section records the
disagreement breakdown and two reporting choices. The verdict scale for disagreement is ordered: reject < modify < accept, since and accept/modify disagreement is less severe than an accept/reject disagreement, treating them as unordered categories would understate the true agreement. Reviewers agree exactly on $77.3\%$ of rules, while they differ by one step on the ordinal scale on $13.1\%$, and differ at the extremes (\texttt{accept}/\texttt{reject}) on $9.6\%$. Per subset the split is $76.0/15.1/8.8$ on Berka, $79.0/10.3/10.7$ on H\&M, $76.8/13.8/9.4$ on agentic rules and $78.1/11.9/10.0$ on guided rules. 
We report AC2 alongside $\alpha$ because the verdict distribution is
concentrated: with $84.8\%$ of ratings in one category, $\alpha$ is deflated by construction, while AC2 is prevalence-robust.


\section{Benchmarking Agenda}
\label{sec:appendix-benchmark}

The experiments in \S\ref{sec:evaluation} focus on rule-generation quality
and diversity. Comprehensive valuation dimensions for the full envisioned system extend beyond these to include: execution correctness (reproducibility of ExecutionTraces from artifact snapshots); ontology grounding quality (semantic role accuracy, confidence calibration); HITL efficiency (accept/modify/reject rates, iterations-to-convergence, reviewer effort proxies); temporal adaptation (portfolio stability under schema drift, MTTD/MTTR); and explanation quality (usefulness of RCA hypotheses and prescriptive suggestions).

As mentioned in the discussion, we thus envision benchmarks that ship not only tables, but the \emph{governance-and-audit context} that the system consumes and produces:
\begin{itemize}
    \item \textbf{Data + evolution timeline}: snapshots/streams of related tables (finance as anchor), with scripted schema/provider/policy evolution and injected drifts/anomalies.
    \item \textbf{Ontology spine}: a typed Dataset/Entity/Attribute/Relationship/Vocabulary ontology linking columns/joins to meaning envelopes (\texttt{constraints}, \texttt{typical\_behavior}, \texttt{risk\_notes}).
    \item \textbf{Governance corpus}: policy/regulatory/dictionary text mapped (at least coarsely) to ontology nodes, enabling provenance links from rules back to intent.
    \item \textbf{Repository event log}: an append-only history of rule versions, HITL decisions, and ExecutionTraces, so adaptation and auditability are testable rather than asserted.
\end{itemize}

We aim, together with the broader community to build these shared benchmark packs and scoring harnesses around this contract, with license-clean datasets, governance corpora, gold rule catalogs, and standard scoring for precision/recall, coverage, stability, and explanation quality.

\section{Rule Taxonomy}
\label{sec:rule-taxonomy}
To characterize what the system actually mines rather than only how well the
rules it mines are received, we manually tagged a randomly drawn sample of
85 AI-mined rules from the Full Agentic mode (25\% of all AI-mined rules) against eight standard
relational and data-quality rule categories. \autoref{tab:rule-taxonomy}
reports the resulting category shares. We provide also representative examples of AI-mined rules for each of the classes present in the table.


\paragraph{Range / statistical}

\begin{quote}
\small
\begin{tabular}{p{0.95\linewidth}}
\ttfamily
\{\\
"rule\_id": "rule:client\_district\_id\_range\_validity",\\
\ \ "rule\_name": "District ID Range Constraint",\\
\ \ "linked\_nodes": ["col:client\_district\_id",\\ "rel:client\_to\_district"],\\
\ \ "logic": \{\\
\ \ \ "intent": "Ensure district\_id falls within the valid administrative range, protecting referential integrity.",\\
\ \ \ "condition": "All non-null district\_id values in client.",\\
\ \ \ "expectation": "Integers between 1 and 77 inclusive.",\\
\ \ \ "supporting\_evidence": "Ontology states integer 1--77; profiling shows min=1, max=77, distinct\_count=77."\\
\ \ \},\\
\ \ "executable\_code": "SELECT * FROM client WHERE\\
\ \ \ \ district\_id $<$ 1 OR district\_id $>$ 77\\
\ \ \ \ OR district\_id IS NULL"\\
\}
\end{tabular}
\end{quote}

\paragraph{Key / uniqueness}
\begin{quote}
\small
\begin{tabular}{p{0.95\linewidth}}
\ttfamily
\{\\
\ \ "rule\_id": "rule:client\_client\_id\_uniqueness",\\
\ \ "rule\_name": "Client ID Uniqueness Constraint",\\
\ \ "linked\_nodes": ["col:client\_client\_id", "tbl:client"],\\
\ \ "logic": \{\\
\ \ \ "intent": "Enforce that client\_id serves as a valid primary key by ensuring each value appears exactly once.",\\
\ \ \ "condition": "All rows where client\_id is present.",\\
\ \ \ "expectation": "Every client\_id unique across all 5369 rows, no duplicates permitted.",\\
\ \ \ "supporting\_evidence": "Profiling shows distinct\_count=5369 equals row\_count=5369, confirming uniqueness."\\
\ \ \},\\
\ \ "executable\_code": "SELECT * FROM client WHERE client\_id IN\\
\ \ \ \ (SELECT client\_id FROM client GROUP BY client\_id\\
\ \ \ \ \ HAVING COUNT(*) $>$ 1)"\\
\}
\end{tabular}
\end{quote}

\paragraph{Categorical domain / vocabulary}

\begin{quote}
\small
\begin{tabular}{p{0.95\linewidth}}
\ttfamily
\{\\
"rule\_id": "rule:card\_type\_vocabulary\_conformance",\\
\ \ "rule\_name": "Card Type Vocabulary Conformance",\\
\ \ "linked\_nodes": ["col:card\_type", "vocab:card\_type\_codes"],\\
\ \ "logic": \{\\
\ \ \ "intent": "Enforce that card type values conform to the controlled vocabulary of valid card tier codes.",\\
\ \ \ "condition": "All rows in card where type is not null.",\\
\ \ \ "expectation": "Every value one of: classic, junior, gold.",\\
\ \ \ "supporting\_evidence": "Profiling shows exactly 3 distinct values: classic (659), junior (145), gold (88), matching vocab:card\_type\_codes."\\
\ \ \},\\
\ \ "executable\_code": "SELECT * FROM card WHERE type NOT IN\\
\ \ \ \ ('classic', 'junior', 'gold')"\\
\}
\end{tabular}
\end{quote}

\paragraph{Syntactic format}

\begin{quote}
\small
\begin{tabular}{p{0.95\linewidth}}
\ttfamily
\{\\
\ \ "rule\_id": "rule:loan\_date\_format\_iso8601",\\
\ \ "rule\_name": "Loan Date ISO 8601 Format Validation",\\
\ \ "linked\_nodes": ["col:loan\_date"],\\
\ \ "logic": \{\\
\ \ \ "intent": "Ensure date values conform to ISO 8601 (YYYY-MM-DD) and are valid calendar dates.",\\
\ \ \ "condition": "All rows in the loan table.",\\
\ \ \ "expectation": "4-digit year, month 01--12, day 01--31, and the combination forms a valid calendar date.",\\
\ \ \ "supporting\_evidence": "Ontology requires fixed-length ISO 8601; profiling shows all 682 values of length 10.0 with std\_length=0.0."\\
\ \ \},\\
\ \ "executable\_code": "SELECT * FROM loan WHERE date !\textasciitilde{}\\
\ \ \ \ '\textasciicircum{}\textbackslash d\{4\}-(0[1-9]$|$1[0-2])-(0[1-9]$|$[12]\textbackslash d$|$3[01])\$' OR date $\neq$ TO\_DATE(date, 'YYYY-MM-DD')::text"\\
\}
\end{tabular}
\end{quote}

\paragraph{Completeness / NOT NULL}

\begin{quote}
\small
\begin{tabular}{p{0.95\linewidth}}
\ttfamily
\{\\
"rule\_id": "rule:trans\_account\_id\_non\_null\_positive",\\
\ \ "rule\_name": "Account ID Non-Null and Positive",\\
\ \ "linked\_nodes": ["col:trans\_account\_id"],\\
\ \ "logic": \{\\
\ \ \ "intent": "Ensure account\_id is always populated and valid as a foreign key reference to accounts.",\\
\ \ \ "condition": "For all rows in the trans table.",\\
\ \ \ "expectation": "account\_id must be non-null and a positive integer ($\geq$ 1).",\\
\ \ \ "supporting\_evidence": "Ontology states account\_id cannot be null and must be a positive integer; profiling confirms 0 nulls, min=1, across all 1{,}056{,}320 rows."\\
\ \ \},\\
\ \ "executable\_code": "SELECT * FROM trans WHERE account\_id\\
\ \ \ \ IS NULL OR account\_id $<$ 1"\\
\}
\end{tabular}
\end{quote}

\paragraph{Functional dependency}

\begin{quote}
\small
\begin{tabular}{p{0.95\linewidth}}
\ttfamily
\{\\
\ \ "rule\_id": "rule:perceived\_colour\_master\_id\_name\_consistency",\\
\ \ "rule\_name": "Perceived Colour Master ID-Name Consistency",\\
\ \ "linked\_nodes": ["col:article\_perceived\_colour\_master\_id",\\
\ \ \ \ \ \ \ \ \ \ \ \ \ \ \ \ \ \ "col:article\_perceived\_colour\_master\_name"],\\
\ \ "logic": \{\\
\ \ \ "intent": "Enforce that each master colour id maps to exactly one master colour name (id $\rightarrow$ name).",\\
\ \ \ "condition": "Every distinct id value, excluding the -1 sentinel for unknown.",\\
\ \ \ "expectation": "Distinct (id, name) pairs equal distinct id values (20); each id maps to a single name.",\\
\ \ \ "supporting\_evidence": "Both columns have exactly 20 distinct values across 105{,}542 rows; matching distinct counts indicate a controlled vocabulary relationship."\\
\ \ \},\\
\ \ "executable\_code": "WITH inconsistent\_ids AS (SELECT id FROM\\
\ \ \ \ article WHERE id $\neq$ -1 GROUP BY id HAVING\\
\ \ \ \ COUNT(DISTINCT name) $>$ 1) SELECT * FROM article\\
\ \ \ \ WHERE id IN (SELECT id FROM inconsistent\_ids)"\\
\}
\end{tabular}
\end{quote}

\paragraph{Inclusion dependency}

\begin{quote}
\small
\begin{tabular}{p{0.95\linewidth}}
\ttfamily
\{\\
\ \ "rule\_id": "rule:account\_district\_referential\_integrity",\\
\ \ "rule\_name": "Account District Referential Integrity",\\
\ \ "linked\_nodes": ["rel:account\_to\_district",\\
\ \ \ \ \ \ \ \ \ \ \ \ \ \ \ \ \ \ "rel:district\_id\_to\_name"],\\
\ \ "logic": \{\\
\ \ \ "intent": "Ensure every account.district\_id references a valid district record in the district table.",\\
\ \ \ "condition": "Every row in the account table.",\\
\ \ \ "expectation": "district\_id must match an existing A1 value in district (1--77 inclusive).",\\
\ \ \ "supporting\_evidence": "Profiling shows 77 distinct district\_id values matching the district table's 77 rows."\\
\ \ \},\\
\ \ "executable\_code": "SELECT * FROM account WHERE district\_id\\
\ \ \ \ IS NULL OR district\_id NOT IN (SELECT A1 FROM district)"\\
\}
\end{tabular}
\end{quote}

\paragraph{Denial constraint}

\begin{quote}
\small
\begin{tabular}{p{0.95\linewidth}}
\ttfamily
\{\\
\ \ "rule\_id": "rule:district\_A5\_A6\_A7\_demographic\_coherence",\\
\ \ "rule\_name": "District A5-A6-A7 Demographic Coherence",\\
\ \ "linked\_nodes": ["col:district\_A5", "col:district\_A6", "col:district\_A7", "rel:demographic\_metrics\_coherence"],\\
\ \ "logic": \{\\
\ \ \ "intent": "Ensure internal consistency among related\\
\ \ \ \ demographic metrics A5, A6 and A7.",\\
\ \ \ "condition": "Every row where A5, A6 and A7 are all non-zero.",\\
\ \ \ "expectation": "Pairwise ratios remain within the bounds implied by observed means (A5:A6:A7 $\approx$ 8:4:1); extreme deviations indicate quality issues.",\\
\ \ \ "supporting\_evidence": "All three columns share similar zero rates (5--6\%); ontology asserts internal coherence."\\
\ \ \},\\
\ \ "executable\_code": "SELECT * FROM district WHERE A5 $>$ 0 AND\\
\ \ \ \ A6 $>$ 0 AND A7 $>$ 0 AND ((A5/A6 $<$ 0.67 OR A5/A6 $>$ 6)\\
\ \ \ \ OR (A5/A7 $<$ 2.57 OR A5/A7 $>$ 23.1)\\
\ \ \ \ OR (A6/A7 $<$ 1.3 OR A6/A7 $>$ 11.7))"\\
\}
\end{tabular}
\end{quote}

\begin{table*}[t]
\small
\centering
\begin{tabular}{p{3.6cm}p{8.2cm}r}
\hline
\textbf{Category} & \textbf{Applied to} & \textbf{Share} \\
\hline
Range / statistical & numeric bounds, thresholds, sign, outlier bounds & 24.2\% \\
Key / uniqueness & primary-key or unique-column assertions & 23.2\% \\
Categorical domain / vocabulary & enumerated valid-value list & 22.1\% \\
Syntactic format & regex, length, type, encoding, calendar-validity of a date string & 10.5\% \\
Completeness / NOT NULL & presence assertions & 9.5\% \\
Functional dependency & $X \to Y$ within a table & 6.3\% \\
Inclusion dependency (IND) & foreign-key / referential integrity across tables & 3.2\% \\
Denial constraint (DC) & forbidden predicate combination not reducible to the categories above & 1.1\% \\
\hline
\end{tabular}
\caption{Taxonomy of AI-mined rule types, tagged on a random 25\% sample
($n=85$) of all AI-mined rules.}
\label{tab:rule-taxonomy}
\end{table*}

\section{Additional Figures}
\label{sec:additional-figs}
Below we provide a series of additional figures and plots to accompany and complement visually the results discussed in \S\ref{sec:results}:

\begin{figure*}[t]
    \centering
    \includegraphics[width=0.95\textwidth]{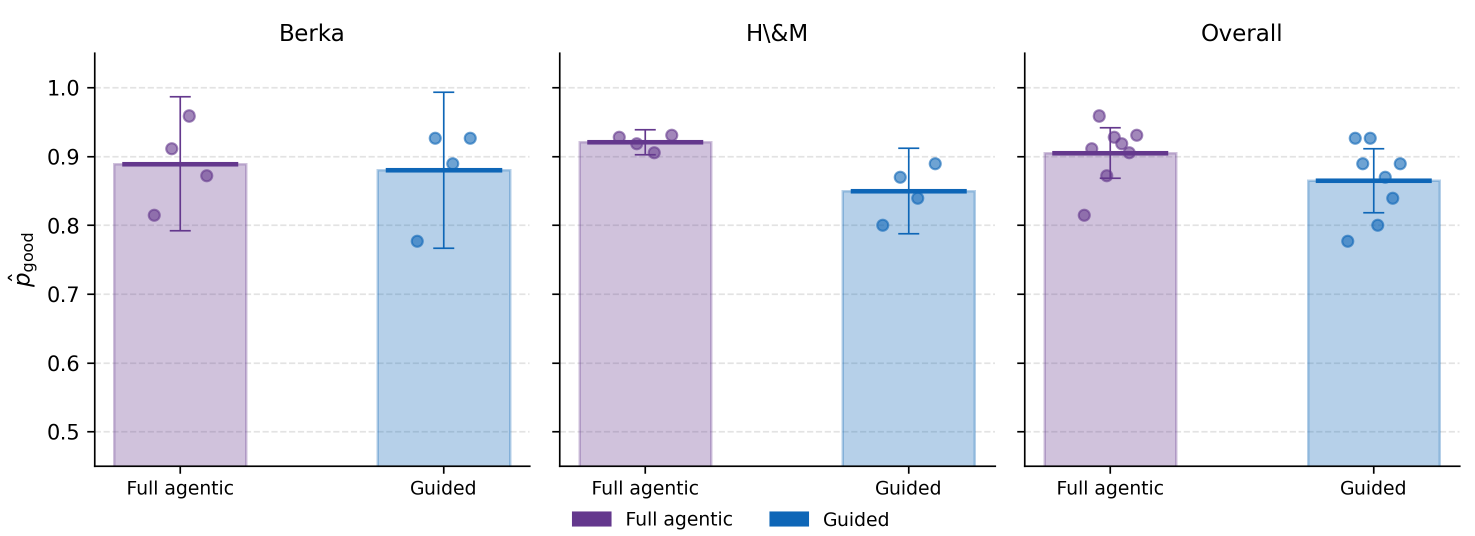}
    \caption{Good-rule rate $\hat{p}_{\mathrm{good}}$ per mode and dataset
    (dots = individual runs, bar = mean, error bars = 95\% $CI$-interval on mean).}
    \label{fig:pgood}
\end{figure*}

\begin{figure*}[t]
    \centering
    \includegraphics[width=0.95\textwidth]{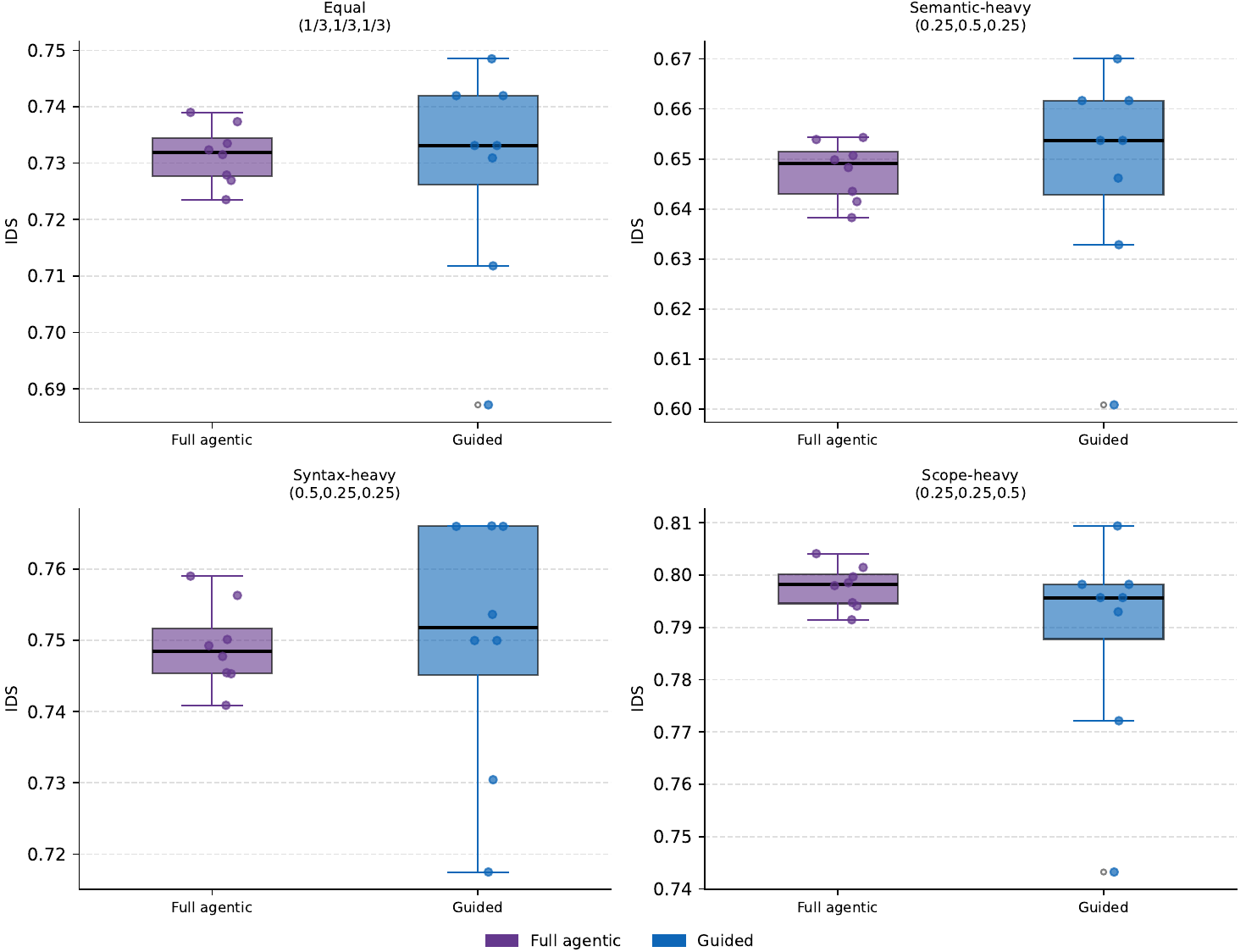}
    \caption{IDS distributions per mode across weight configurations
    ($n_1 = n_2 = 8$ runs per mode, pre-review rules).}
    \label{fig:ids_dist}
\end{figure*}

\begin{table*}[t]
\small
\centering
\setlength{\tabcolsep}{5pt}
\begin{tabular}{llccccc}
\hline
\textbf{Dataset} & \textbf{Mode} & $N_{\mathrm{rules}}$ & $N_{\mathrm{ratings}}$ & Accept & Modify & Reject \\
\hline
\multirow{2}{*}{Berka}
  & Full agentic & 216 & 432 & 363 (84.0\%) & 42 (9.7\%) & 27 (6.2\%) \\
  & Guided    & 101 & 202 & 168 (83.2\%) & 18 (8.9\%) & 16 (7.9\%) \\
\hline
\multirow{2}{*}{H\&M}
  & Full agentic & 124 & 248 & 222 (89.5\%) & 13 (5.2\%) & 13 (5.2\%) \\
  & Guided    & 100 & 200 & 164 (82.0\%) & 12 (6.0\%) & 24 (12.0\%) \\
\hline
\multirow{2}{*}{Overall}
  & Full agentic & 340 & 680 & 585 (86.0\%) & 55 (8.1\%) & 40 (5.9\%) \\
  & Guided    & 201 & 402 & 332 (82.6\%) & 30 (7.5\%) & 40 (10.0\%) \\
\hline
\end{tabular}
\caption{HITL review outcomes per mode and dataset (pooled across $8$ runs).
$N_{\mathrm{rules}}$: rules proposed; $N_{\mathrm{ratings}} =
2N_{\mathrm{rules}}$, since the blinded block design has every rule rated by
exactly two reviewers who did not author it (\S\ref{sec:review-protocol}). Accept/Modify/Reject are counts over ratings, with row percentages. Good-rule rates are reported in \autoref{tab:rq1-summary} as the mean across good rule rates of reviewer runs. The Full Agentic mode proposes more rules than Guided because the orchestrator enforces a per-column coverage criterion, whereas a human expert might target a narrower set of constraints. The matched-scope control in \S\ref{sec:results} re-runs the RQ1 analysis with this volume asymmetry removed.}
\label{tab:hitl-outcomes}
\end{table*}

\end{document}